\documentclass{ifacconf}

\usepackage{graphicx}      
\usepackage{natbib}        
\usepackage{amsmath}
\usepackage{amssymb}
\usepackage{subfigure}
\usepackage{cuted}
\usepackage{caption}
\newtheorem{assumption}{Assumption}
\begin{document}
\begin{frontmatter}

\title{Impact-Aware Model Predictive Control for UAV Landing on a Heaving Platform}

\author[First]{Jess Stephenson} 
\author[First]{Melissa Greeff}

\address[First]{Robora Lab, Ingenuity Labs Research Institute, Department of Electrical and Computer Engineering, Queen's University, Kingston, Canada, (e-mail: jess.stephenson, melissa.greeff, @queensu.ca)}

\begin{abstract}            
Landing UAVs on heaving marine platforms is challenging because relative vertical motion can generate large impact forces and cause rebound on touchdown. To address this, we develop an impact-aware Model Predictive Control (MPC) framework that models landing as a velocity-level rigid-body impact governed by Newton’s restitution law. We embed this as a linear complementarity problem (LCP) within the MPC dynamics to predict the discontinuous post-impact velocity and suppress rebound. In simulation, restitution-aware prediction reduces pre-impact relative velocity and improves landing robustness. Experiments on a heaving-deck testbed show an 86.2\% reduction in post-impact deflection compared to a tracking MPC.
\end{abstract}

\end{frontmatter}
\section{Introduction} 
Unmanned Aerial Vehicles (UAVs) are increasingly deployed in marine disaster scenarios, including search-and-rescue missions, evaluating fires on offshore oil platforms, and rescuing drowning victims (\cite{yeong_review_2015}, \cite{lyu_unmanned_2023}). However, providing persistent aerial support in open-water environments remains a significant challenge due to limited flight endurance and a lack of safe landing or recharging opportunities offshore. Enabling UAVs to land safely and reliably on Uncrewed Surface Vessels (USVs) is a critical step towards enhancing the longevity, autonomy, and operational range of marine disaster management operations. This capability would allow UAVs to continuously recharge, enabling remote deployment without the need for a manned support vessel or a static base station.

Roll, pitch, and especially heave pose the most significant risks during landing, while surge, sway, and yaw generally induce only minor disturbances at touchdown (\cite{ross_autonomous_2021}). In wave-excited marine environments, USVs exhibit heave motion that drives vertically varying relative motion between the UAV and the deck. This relative vertical motion—particularly when the UAV descends as the platform ascends—can generate large, uneven impact forces if not anticipated and controlled (\cite{xia2022landing}). Previous work has focused mainly on tilt, developing strategies for landing near-zero-tilt (\cite{gupta2022landing}, \cite{stephenson_time}), attitude alignment (\cite{persson2021model}), and landing on static inclined surfaces (\cite{Panagiotis2015}). In contrast, far less attention has been given to heave motion, despite its critical role in enabling safe, controlled landings on dynamically moving surfaces.
\begin{figure}
\centering
\includegraphics[width=0.47\textwidth,trim={0cm 0.25cm 11.3cm 0.0cm},clip]{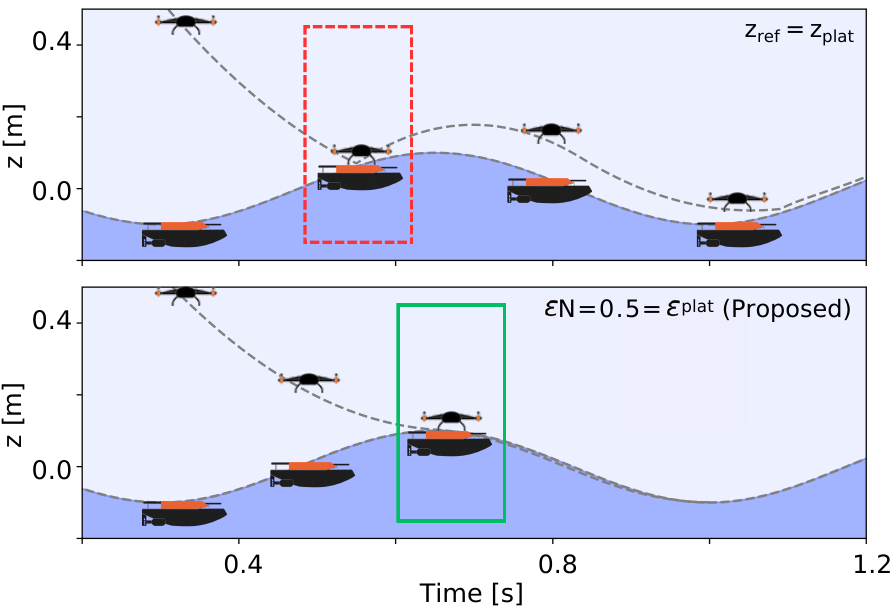}
\caption{Simulated landing trajectories for the baseline (top) and proposed impact-aware MPC (bottom) on a heaving deck. The baseline rebounds due to high pre-impact relative velocity, while the proposed method mitigates the impact and enables a safe landing.}
 \label{intro_fig}
\end{figure}



\section{Related Work}
Several mechanical approaches have been proposed for landing multirotor UAVs on heaving vessels. These include augmenting the deck or UAV with specialized landing gear (\cite{leon_rotorcraft_2021}), using vision-based guidance systems (\cite{sanchez-lopez_approach_2014}), or tethering the vehicle to the ship (\cite{alarcon_precise}). While such methods demonstrate successful landings in simulation, controlled experiments, or hardware-in-the-loop evaluations, each carries notable limitations: mechanical aids reduce interoperability across platforms (\cite{abujoub_methodologies_2020}), vision systems degrade in adverse weather or low-light conditions (\cite{ross_zero_2019}), and tethers fundamentally restrict operational range.

Existing control-based approaches primarily use reference-tracking controllers, similar to the baseline in Fig. \ref{intro_fig}, to regulate the relative motion between the UAV and the vessel. Methods such as \cite{hervas_automatic_2014, marconi_autonomous_2002, xuan-mung_autonomous_2020} track a time-varying reference derived from the vessel’s heave motion, aiming to drive the relative altitude error to zero. These works, therefore, address the problem of estimating and following an irregular, indirectly measured trajectory. However, minimizing vertical tracking error alone does not guarantee a safe landing: even with perfect knowledge of the platform trajectory, the UAV can still contact the deck with substantial relative impact velocity, exposing it to harmful impulsive forces that accelerate hardware fatigue or cause catastrophic damage.

To mitigate this, \cite{abujoub_methodologies_2020} proposes an Active Heave Compensation (AHC) strategy that searches over candidate landing trajectories to maintain impact velocity below a safe threshold. In contrast, we embed an explicit impact model directly within the Model Predictive Control (MPC) prediction loop, enabling the controller to reason about and reduce future impact dynamics in real time rather than selecting a precomputed trajectory.


Rigid-body collisions are commonly modeled as impulsive events, as in \cite{posa_direct_2014}, where instantaneous forces produce a discontinuous jump in velocity. To handle these discontinuities in continuous-time control frameworks, many works enforce contact through complementarity conditions—a standard tool in rigid-body interaction domains such as quadruped locomotion (\cite{lecleach}) and cable-suspended payload control (\cite{foehn_fast_2017}). These conditions encode that bodies are either separated with zero contact force or in contact with a nonnegative force, but never both. This formulation integrates naturally with MPC, where complementarity constraints can be embedded directly into the system dynamics to regulate contact events.

Motivated by the analogy between a quadruped foot strike and a multirotor touchdown, we adopt complementarity-based contact dynamics to model UAV landings. Prior complementarity-based MPC work for aerial interaction (\cite{fuser_linear_2025}) considers only static platforms and uses position-level complementarity, effectively assuming a perfectly inelastic collision where post-impact normal velocity is forced to zero. While this prevents penetration, it does not capture how velocity should evolve at impact. Velocity-level formulations instead constrain the normal relative velocity and admit discontinuous updates governed by impact laws such as Newton’s restitution (used in \cite{banerjee_oblique_2016}). This distinction is critical for UAV landings, where post-impact velocity determines whether the vehicle settles or rebounds. To address this gap, we incorporate Newton’s restitution law into MPC, enabling the prediction of discontinuous post-impact behavior necessary for safe landings on a heaving platform. The key contributions of this paper are:
\begin{itemize}
    \item We present a restitution-aware MPC framework that embeds a relative-motion Linear Complementarity Problem (LCP) for UAV landing on a heaving platform. By modeling touchdown as a rigid-body impact governed by Newton’s restitution law and expressed in terms of UAV–deck relative velocity, the controller anticipates discontinuous dynamics and actively suppresses rebound.
    \item We show that overestimating the restitution coefficient in simulation leads the controller to predict a harsher rebound than reality. This causes a more conservative approach behaviour and produces softer, safer landings in practice—highlighting a key safety advantage of embedding restitution-based impact prediction directly within the MPC loop.
    \item We validate the proposed method on a physical heaving platform across four regimes mimicking different sea states, and observe that it reduces post-impact deflection by 86.2\% compared to a standard tracking MPC, resulting in a substantial increase in landing success.
\end{itemize}

\section{Problem Statement}
We consider a multirotor UAV with state $\mathbf{x} = [\mathbf{q}, \dot{\mathbf{q}}]^\top$, where $\mathbf{q}$ and $\dot{\mathbf{q}}$ are the generalized coordinates and corresponding velocities respectively. Before contact, the vehicle evolves according to the discrete-time free-flight dynamics:
\begin{equation}
    \mathbf{x}_{k+1} = f_{\mathrm{free}}(\mathbf{x}_k, \mathbf{u}_k),
\label{eq_free}
\end{equation}
where $\mathbf{u}_k$ is the control input subject to actuator limits $\mathbf{u}_k \in \mathcal{U}$, and $\mathbf{x}_k \in \mathcal{X}$ encodes state constraints. The objective is to design a controller that guides the UAV to land safely on a vertically heaving platform whose motion induces significant, time-varying relative velocity during touchdown. We adopt two simplifying assumptions consistent with our landing scenario:
\begin{assumption}
The deck's vertical position $z_{\mathrm{plat}}(t)$ is known at each time step (either measured or predicted). 
    \label{a0}
\end{assumption}
\begin{assumption} 
The contact between the multirotor and the deck is modeled as a 
single, frictionless point contact at the UAV’s center of mass. Tangential 
contact forces and frictional moments are neglected, so that only the 
normal reaction force $f_N \ge 0$ contributes to the impact dynamics.
\label{a1}
\end{assumption}
These assumptions are justified by the operating regime of interest: the deck is approximately horizontal (we ignore pitch and roll), and the multirotor approaches with negligible horizontal velocity, so relative motion at touchdown is 
dominated by the normal direction. Safe landing is therefore determined 
primarily by the normal relative velocity and the resulting normal impulse, 
while tangential forces do not play a significant role in rebound or post–impact 
behavior. Modeling only the normal contact direction is thus sufficient for 
capturing the impact dynamics relevant to our control design.

We define the normal gap function $g_N(\mathbf{q}, t)\in\mathbb{R}^m$ and relative normal velocity $\dot{{g}}_N(\mathbf{q}, \dot{\mathbf{q}}, t)\in\mathbb{R}^m$ as:
\begin{align}
    g_N(\mathbf{q}, t) &= {z} - z_{\mathrm{plat}}(t) \label{gN}, \\
    \dot{g}_N(\mathbf{q}, \dot{\mathbf{q}}, t) &= \dot{z} - \dot z_{\mathrm{plat}}(t),\label{g_dotN}
\end{align}
where $z$ is the UAV's altitude and $\dot{z}$ its descent velocity. A touchdown event occurs at an unknown time $t_k$ when the vertical distance between the multirotor and platform is zero,  $g_{N} = 0$, and the UAV is descending $\dot g_{N} < 0$, producing a discontinuous jump in velocity described by an impact map:
\begin{equation}
    \mathbf{x}(t_k^+) = \Delta\big(\mathbf{x}(t_k^-)\big).
\label{eq_impactmap}
\end{equation}

The control objective is to compute the input $\mathbf{u}_k$ at each time step $k$ that:
\begin{itemize}
    \item limits the pre-impact relative velocity $\dot{g}_N^-$ to prevent rebound or unsafe contact forces,
    \item achieves desirable post-impact behavior--i.e., limited rebound and rapid settling onto the deck--that we verify empirically through simulation and hardware experiments, and
    \item satisfies the state and input constraints $\mathbf{x}_k \in \mathcal{X}$ and $\mathbf{u}_k \in \mathcal{U}$ for all $k$.
\end{itemize}
  
In summary, the problem is to construct a control policy that steers the UAV from its initial state to a safe, stable touchdown on a heaving platform--explicitly accounting for contact constraints, rigid-body impact dynamics, and the relative motion induced by deck's heave.

\section{Background}
We begin by introducing the free-flight multirotor dynamics (\ref{eq_free}), which serve as the pre-contact baseline model. We then review commonly used rigid-body impact models (\ref{eq_impactmap}), highlighting why standard formulations are insufficient for UAV landings on a vertically moving deck. These limitations motivate the restitution-based impact model leveraged within the control methodology of Section \ref{sec_method}.

\subsection{Multirotor Dynamics}
\begin{assumption} The multirotor’s motion is constrained to a plane, reducing 
the problem to a two-dimensional setting.
\label{a2}
\end{assumption}
Let $\{W\}$ denote an inertial frame and $\{B\}$ a body-fixed frame at the 
multirotor’s center of mass. The multirotor is modeled as a rigid body with 
configuration $\mathbf{q} = [x, z, \theta]^\top$ comprising of position $x,z$ and pitch $\theta$, and mass–inertia matrix 
$\mathbf{M}=\mathrm{diag}(m_r, m_r, I_r)$. The continuous dynamics are:
\begin{equation}
    \begin{aligned}
\mathbf{M}\ddot{\mathbf{q}}=m_rg\mathbf{e}_3 + \mathbf{B}(\mathbf{q})\mathbf{u} + \mathbf{J}_N(\mathbf{q})^\top f_N,
    \end{aligned}
    \label{multirotor}
\end{equation}
where $\mathbf{u}=[T, \tau_\theta]^\top$ consists of total thrust and pitch torque, 
and $\mathbf{B}(\mathbf{q})$ maps control inputs to generalized forces.
To model contact, we use only the \emph{normal} Jacobian $J_N(\mathbf{q})\in\mathbb{R}^{1\times 3}$, which maps generalized coordinates to the normal contact coordinate. It is obtained as the gradient of the gap 
function $g_N(\mathbf{q},t)$ with respect to $\mathbf{q}$. The internal or contact force therefore reduces to a single normal reaction $f_N \ge 0$.

Similar to \cite{fuser_linear_2025}, we can discretize the continuous multirotor dynamics (\ref{multirotor}) with a semi-implicit Euler scheme with time step $\Delta t$:
\begin{equation}
\begin{aligned}
\dot{\mathbf{q}}_{k+1}&=\dot{\mathbf{q}}_k+\mathbf{M}^{-1}(-m_rg\mathbf{e_3}+\mathbf{B}_k\mathbf{u}_k+\mathbf{J}_{N,k}^\top f_{N,k+1})\Delta t\\
\mathbf{q}_{k+1}&=\mathbf{q}_k+\Delta t\;\dot{\mathbf{q}}_{k+1}.
\end{aligned}
\label{semi}
\end{equation}

In the time-stepping formulation in (\ref{semi}), the continuous-time normal contact force $f_N$ is represented by the normal impulse $\lambda_{N,k+1} = f_{N,k+1}\Delta t$, which governs the instantaneous velocity jump at impact.

In the absence of contact forces, the evolution in (\ref{semi}) reduces to the free dynamics:
\begin{equation}
\dot{\mathbf{q}}_{k+1}^\text{free}=\dot{\mathbf{q}}_k+\mathbf{M}^{-1}(-m_rg\mathbf{e_3}+\mathbf{B}_k\mathbf{u}_k)\Delta t.
    \label{q_free1}
\end{equation}

\subsection{Complementarity-Based Contact Dynamics}\label{backgroundD}

To prevent interpenetration between the multirotor and the deck, one approach is to
enforce a unilateral constraint on the gap function $g_N(\mathbf q,t)$
defined in (\ref{gN}), which measures the signed distance along the surface
normal between the multirotor and the deck. A touchdown event occurs
when $g_N(\mathbf q,t)=0$ and the bodies exhibit closing motion $\dot g_N<0$
as defined in (\ref{g_dotN}). Before contact ($g_N>0$), the UAV evolves under
the free-flight dynamics in (\ref{q_free1}). At contact, however, rigid-body
collisions produce a discontinuous jump in velocity, requiring a
nonsmooth model of impact dynamics.

Signorini’s law classically describes unilateral (i.e., the surface can push back but never pull the object toward it) contact
(\cite{moreau_unilateral_1988}; \cite{signorini1933sopra}), which couples the normal gap $g_N$ and
normal reaction force $f_N$ via the position-level complementarity
condition
\begin{equation}
  0 \le f_N \;\perp\; g_N \ge 0.
  \label{eq:signorini_position}
\end{equation}

This encodes that either (i) the bodies are separated with
$g_N>0$ and $f_N=0$, or (ii) they are in contact with $g_N=0$ and
a nonnegative reaction force $f_N \ge 0$ acts to prevent penetration.

Differentiating the gap
yields the normal relative velocity $\dot g_N = \tfrac{d}{dt}g_N(\mathbf q,t)$,
which leads to the velocity-level form (\cite{moreau_unilateral_1988})
\begin{equation}
    0 \le f_N \;\perp\; \dot g_N \ge 0,
    \label{eq:signorini_vel}
\end{equation}
enforced only at active contacts ($g_N = 0$). In smoothly deformable or
perfectly inelastic rigid-body contacts, \eqref{eq:signorini_position} and
\eqref{eq:signorini_vel} are equivalent (\cite{jean_non-smooth_1999}).

While position- and velocity-level Signorini conditions coincide for
smoothly deformable bodies or for perfectly inelastic rigid-body
contacts, neither assumption is suitable for our landing scenario.
Smoothly deformable models presuppose continuous force–deformation
relationships and preclude instantaneous velocity jumps, which is
incompatible with the impulsive, rigid-body impacts observed when a
multirotor contacts a hard deck. Conversely, perfectly inelastic
rigid-body contacts force the post-impact normal velocity to zero, eliminating any possibility of rebound. This
oversimplifies the UAV–deck interaction, as real heave-induced
collisions exhibit nonzero restitution and post-impact
velocity directly determines whether the UAV settles or
rebounds. Accurately capturing this behavior, therefore, requires an
explicit restitution-based impact model rather than either of these
limiting assumptions.

\section{Methodology}
\label{sec_method}
Our proposed methodology integrates a nonsmooth rigid-body impact model into an MPC framework for UAV landing. Section \ref{sect:metho_1} introduces the velocity-level restitution model used to describe the instantaneous jump in normal velocity that occurs when the multirotor comes into contact with the deck. This
restitution law provides the closure needed to uniquely determine the post-impact velocity in (imperfectly) inelastic collisions. To embed this impact behavior within a time-stepping scheme, in Section \ref{LCP_sect} we reformulate the
resulting complementarity condition as a Linear Complementarity Problem (LCP), enabling us to compute the normal impact impulse that enforces the restitution law while preventing penetration. In Section \ref{sect:metho_3} the MPC then uses this
LCP-based update within the dynamics constraints, and is further augmented with a restitution residual cost that improves anticipatory behaviour as the vehicle approaches contact. Together, these elements allow the controller to reason about discontinuous impact dynamics and produce safer landing trajectories on a heaving deck.

\subsection{Velocity-Level Restitution for Rigid-Body Impact} \label{sect:metho_1}

For our rigid bodies (multirotor and platform deck) undergoing (imperfectly) inelastic impacts, the Signorini conditions alone do not determine the post-impact normal velocity, since rigid-body collisions generate an instantaneous jump in velocity and
impulsive contact forces, see \cite{jean_non-smooth_1999}. To obtain a well-posed impact update, we instead impose Newton’s restitution law in the normal direction,
\begin{equation}
\dot g_N^{+} = -\epsilon_N\,\dot g_N^{-}, \qquad
\nu_N = \dot g_N^{+} + \epsilon_N \dot g_N^{-},
\label{eq:newton_method}
\end{equation}
where $\dot g_N^{-}$ and $\dot g_N^{+}$ denote the pre- and post-impact
normal relative velocities, and $\epsilon_N\in[0,1]$ is the coefficient of
restitution. The residual $\nu_N$ measures adherence to the restitution
law, with $\nu_N = 0$ if and only if the post-impact velocity satisfies
$\dot g_N^{+} = -\epsilon_N \dot g_N^{-}$. Notably, the velocity-level
Signorini condition \eqref{eq:signorini_vel} corresponds to the special case $\epsilon_N = 0$,
i.e., a perfectly inelastic collision.

To incorporate restitution into the time-stepping dynamics in (\ref{semi}) and (\ref{q_free1}), we enforce the velocity-level complementarity condition
\begin{equation}
0 \;\le \lambda_N \;\perp\; \nu_N \ge 0,
\label{eq:newt_comp_method}
\end{equation}
which activates the normal impulse $\lambda_N$ only when the contact is active and
closing ($g_N = 0$, $\dot g_N^- < 0$). This coupling ensures that a
nonnegative normal impulse enforces the restitution law at impact, while
zero impulse permits separation. This formulation provides a complete
nonsmooth model of the impact update and serves as the basis for the
time-stepping LCP in the MPC dynamics.


\subsection{Linear Complementarity Problem (LCP)} \label{LCP_sect}

To implement the restitution-based complementarity condition (\ref{eq:newt_comp_method}) in a time-stepping algorithm, we reformulate it as an LCP (\cite{anitescu1997formulating, stewart2000implicit}). The standard LCP in \cite{cottle_linear_2009} seeks vectors
$\mathbf{w}$ and $\mathbf{z}$ satisfying
\begin{equation}
\mathbf{w} = \mathbf{Rz} + \mathbf{u}, \qquad
\mathbf{w} \ge 0, \;\; \mathbf{z} \ge 0, \;\; \mathbf{w}^\top \mathbf{z} = 0,
\end{equation}
where $\mathbf{R}$ and $\mathbf{u}$ are known, $\mathbf{z}$ is the
decision variable (e.g., normal impulse $\lambda_N$), and $\mathbf{w}$
represents the associated contact quantity (e.g., restitution residual $\nu_N$).
In this form, the unilateral contact condition
$0 \le \mathbf{z} \perp \mathbf{w} \ge 0$ becomes a standard LCP
solvable using Lemke’s algorithm (\cite{lemke_dual_1954}).

In the time-stepping scheme in (\ref{semi}), the normal contact force $f_N$ is represented by the normal contact impulse $\lambda_N = f_N \Delta t$, which governs the instantaneous velocity jump at impact and serves as the decision variable in the LCP. With this notation, the semi-implicit time-stepping update of the UAV velocity becomes
\begin{equation}
\dot{\mathbf q}_{k+1}
=
\dot{\mathbf q}^{\mathrm{free}}_{k+1}
+
\mathbf M^{-1} \mathbf{J}_{N}^\top \lambda_{N,k+1}.
\label{eq:semi_implicit_update}
\end{equation}
In our case, the normal Jacobian $\mathbf{J}_{N}$ is constant. To incorporate restitution into the time-stepping dynamics, we work with
the relative normal velocity defined by $\dot g_N = \mathbf{J}_{N}\dot{\mathbf q} - \dot z_{\mathrm{plat}}(t),$ which captures the relative motion between the UAV and the heaving deck. Using the semi-implicit update (\ref{eq:semi_implicit_update}), when the UAV comes into contact with the deck, the normal relative velocity post-impact $\dot g^+_{N}$ is determined as:
\begin{equation}
\dot g^+_{N} = \dot g_{N,k+1}
=
\mathbf{J}_{N}\dot{\mathbf q}_{k+1}
-
\dot z_{\mathrm{plat},k+1}.
\label{eq:relative_vel_next}
\end{equation}
Likewise, the pre-impact normal relative velocity  $\dot g^-_{N}$ is
\begin{equation}
\dot g^-_{N} = \dot g_{N,k}
=
\mathbf{J}_{N}\dot{\mathbf q}_{k}
-
\dot z_{\mathrm{plat},k}.
\label{eq:relative_vel_prev}
\end{equation}

Substituting \eqref{eq:semi_implicit_update} into 
\eqref{eq:relative_vel_next} gives
\begin{equation}
\dot g_{N,k+1}
=
J_{N,k}\dot{\mathbf q}^{\mathrm{free}}_{k+1}
-
\dot z_{\mathrm{plat},k+1}
+
\mathbf{J}_{N}\mathbf M^{-1}\mathbf{J}_{N}^\top \lambda_{N,k+1}.
\label{eq:relative_vel_affine}
\end{equation}

Substituting these expressions into Newton’s restitution law (\ref{eq:newton_method}), gives the restitution residual in affine form:
\begin{equation}
\begin{aligned}
\nu_N
&=
\mathbf{J}_{N}\mathbf M^{-1} \mathbf{J}_{N}^\top \lambda_{N,k+1} \\
&\quad +
\Big(
    \mathbf{J}_{N}\dot{\mathbf q}^{\mathrm{free}}_{k+1}
    - \dot z_{\mathrm{plat},k+1}
\Big)
+
\epsilon_N
\Big(
    \mathbf{J}_{N}\dot{\mathbf q}_{k}
    - \dot z_{\mathrm{plat},k}
\Big).
\end{aligned}
\label{eq:nu_full_method}
\end{equation}

\begin{assumption}[Quasi-constant deck velocity]
Over a single timestep $[t_k,t_{k+1}]$, the platform heave velocity varies
negligibly, so that $\dot z_{\mathrm{plat},k+1} \approx \dot z_{\mathrm{plat},k}$.
\end{assumption}

Under this assumption, the restitution residual \eqref{eq:nu_full_method}
simplifies to
\begin{equation}
\nu_N 
\approx
\mathbf{J}_{N}\mathbf M^{-1} \mathbf{J}_{N}^\top \lambda_{N,k+1}
+
(1+\epsilon_N)\big(\mathbf{J}_{N}\dot{\mathbf q}_k - \dot z_{\mathrm{plat},k}\big),
\label{eq:nu_simplified}
\end{equation}
which has the affine LCP form $\mathbf w = \mathbf R \mathbf z + \mathbf u,$
with
\[
\mathbf w = \nu_N,\qquad
\mathbf z = \lambda_{N,k+1},\qquad
\mathbf R = \mathbf{J}_{N}\mathbf M^{-1} \mathbf{J}_{N}^\top,
\]
and $\mathbf u = (1+\epsilon_N)\big(\mathbf{J}_{N}\dot{\mathbf q}_k - \dot z_{\mathrm{plat},k}\big)$. This simplified affine term is used in the time-stepping LCP to compute the normal impact impulse within the MPC prediction horizon. 

The resulting LCP formulation is built on two key components. First, the normal gap and its time derivative are expressed in terms of the relative motion between the UAV and the deck, rather than the UAV’s motion alone. This allows the contact model to capture the platform’s heave dynamics directly, in contrast to stationary-surface formulations such as those used in prior work (e.g., \cite{fuser_linear_2025}). As a result, the normal impact impulse computed by the LCP naturally accounts for a vertically moving landing surface. Second, the LCP incorporates Newton’s restitution coefficient, which can be selected to reflect the mechanical design of the deck--ranging from soft, energy-absorbing materials to more rigid structures. By combining relative kinematics with a tunable restitution law, the LCP provides a model of impact that can be embedded within the MPC prediction loop.

\begin{figure}[t]
\centering
\includegraphics[width=0.47\textwidth,trim={0.4cm 0.48cm 0.3cm 0.3cm},clip]{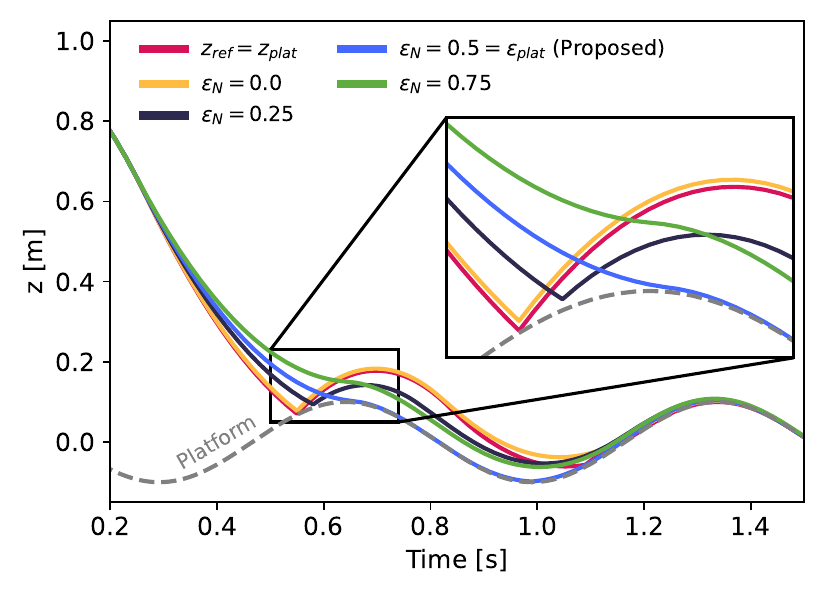}
 \caption{Landing trajectories for all five simulated MPC-based strategies. When $\epsilon_N<\epsilon_\text{plat}$ (yellow, black), the UAV impacts the platform with high pre-impact relative velocity and rebounds off the surface. For $\epsilon_N\geq\epsilon_\text{plat}$ (blue, green), the UAV thrusts to reduce downward velocity before reaching the platform, eliminating the rebounding behaviour at impact.}
 \label{full_sim}
\end{figure}

\begin{figure}[t]
 \centering
 \subfigure[$\epsilon_N=0.0$]{
      \centering
      \label{sim1}
      \includegraphics[width=0.8\columnwidth,trim={0.43cm 0.48cm 0.46cm 0.36cm},clip]{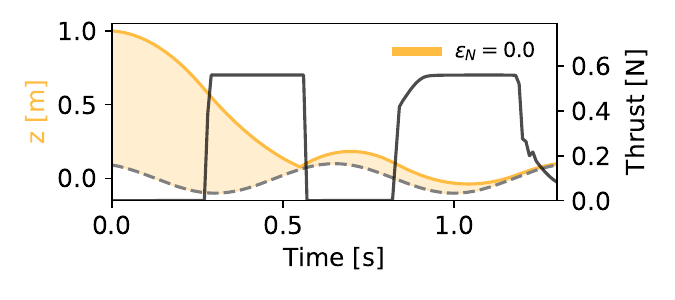}
      }
 \subfigure[$\epsilon_N=0.25$]{
      \centering
      \label{sim2}
      \includegraphics[width=0.8\columnwidth,trim={0.43cm 0.48cm 0.46cm 0.0cm},clip]{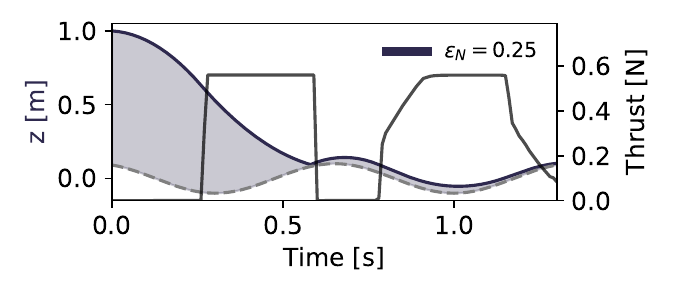}
 }
\subfigure[$\epsilon_N=0.5=\epsilon_\text{plat}$ (Proposed)]{
      \centering
      \label{sim3}
      \includegraphics[width=0.8\columnwidth,trim={0.43cm 0.48cm 0.46cm 0.0cm},clip]{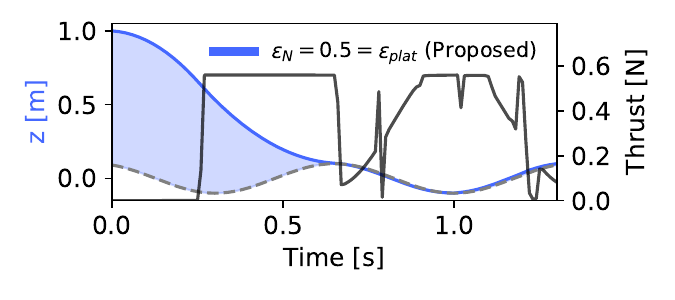}
 } 
 \subfigure[$\epsilon_N=0.75$]{
      \centering
      \label{sim4}
      \includegraphics[width=0.8\columnwidth,trim={0.43cm 0.48cm 0.46cm 0.0cm},clip]{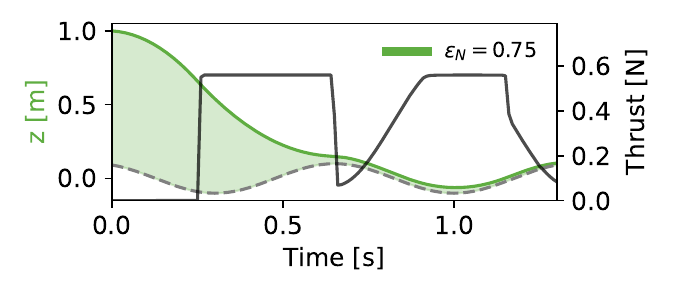}
 } 
 \caption{Individual trajectories and associated commanded thrust profiles for four simulated strategies. In (c) our proposed approach achieves the lowest mean absolute error (in units of meters) between the UAV and the platform, this is seen in the highlighted area under the curve. We observe both pre- and post-contact error for $\epsilon_N<\epsilon_\text{plat}$ strategies ((a) and (b)) and $\epsilon_N>\epsilon_\text{plat}$ (d).}
 \label{partial_sim}
\end{figure}

\subsection{LCP-Integrated Model Predictive Control} \label{sect:metho_3}

The LCP \eqref{eq:nu_simplified} embeds the restitution-based 
complementarity condition \eqref{eq:newt_comp_method} directly into the MPC 
via the time-stepping dynamics, ensuring that the post-impact normal velocity 
satisfies Newton’s restitution law with coefficient $\epsilon_N$. At each 
prediction step, the MPC solves the constrained problem
\begin{equation}
\begin{aligned}
    \min_{\mathbf{x},\,\mathbf{u}} \quad & J(\mathbf{x},\mathbf{u}) \\
    \text{s.t.} \quad 
    & \mathbf{x}_{k+1} = \boldsymbol{f}(\mathbf{x}_k,\mathbf{u}_k), \\
    & \mathbf{x}_k \in \mathcal{X}, \\
    & \mathbf{u}_k \in \mathcal{U}, \\
    & \mathbf{x}_0 = \mathbf{x}_{\mathrm{init}},
\end{aligned}
\label{eq:mpc_problem}
\end{equation}
where $\boldsymbol{f}(\cdot)$ is the semi-implicit update 
\eqref{eq:semi_implicit_update}. The normal impulse $\lambda_{N,k+1}$ is obtained 
by solving the scalar (in our case) LCP $w = Rz + u$, which admits the closed-form solution $z = \max\!\left(0,\,-\frac{u}{R}\right)$ (\cite{van_bokhoven_explicit_1999}). Thus, the complementarity condition is enforced implicitly through the 
dynamics without introducing an additional decision variable. This follows a nested structure similar to \cite{fuser_linear_2025}, where the LCP is the inner problem that solves for $\lambda_{N,k+1}$, while the outer problem \eqref{eq:mpc_problem} is solved by the MPC using the result of the inner problem in $\boldsymbol{f}(\cdot)$.

Because the LCP is only active at impact, the optimizer receives no gradient 
information beforehand. To supply anticipatory guidance, we introduce a 
restitution cost
\begin{equation}
J_R = \sum_{k=0}^{N-1} 
\|\nu_{N,k}\|_{W}^{2}
\
\end{equation}
which penalizes the velocity-level residual $\nu_N$ over predictive horizon $N$ with nonnegative weight $W \geq 0$. Although $\nu_N \neq 0$ 
in free flight, evaluating this residual as if impact were to occur at the 
current timestep provides a meaningful sensitivity signal: it reflects the 
hypothetical post-impact velocity implied by the current approach speed. 
Penalizing this “virtual” post-impact velocity shapes the optimizer’s behaviour 
before contact, discouraging high closing speeds even when the complementarity 
constraint is inactive. Combined with the embedded LCP update, this encourages 
reduced pre-impact relative velocity and yields softer, safer landings on a 
heaving platform.

The MPC cost $J(\cdot)$ in (\ref{eq:mpc_problem}) is therefore composed of two parts: the general tracking cost $J_T(\cdot)$, and a restitution cost $J_R(\cdot)$, i.e., $J(\cdot)=J_T(\cdot)+J_R(\cdot)$. We use a standard quadratic tracking cost:
\begin{equation}
J_T = 
\sum_{k=0}^{N-1} 
\left(
\|\mathbf{x}_k - \mathbf{x}_{\mathrm{ref},k}\|_{\mathbf{Q}}^2 
+
\|\mathbf{u}_k\|_{\mathbf{R}}^2
\right),
\end{equation}
where $\mathbf{x}_{\mathrm{ref},k}$ is the landing deck's state (position and velocities) at time step $k$, and $\mathbf{Q} \succeq 0$, $\mathbf{R} \succ 0$ are the state error and input weighting matrices respectively.

\begin{figure}[t]
\centering
\includegraphics[width=0.39\textwidth,trim={0.4cm 0.55cm 0.4cm 0.4cm},clip]{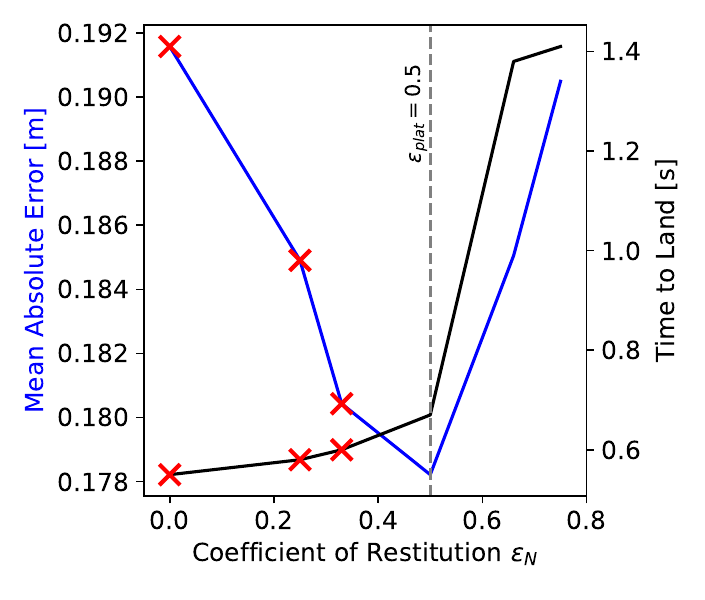}
 \caption{We see, in simulation, that $\epsilon_N<\epsilon_\text{plat}$ reduces pre-contact error by reducing time to land, however all of these attempts rebound off the surface due to high pre-impact approach velocity. Deflection at impact is an unsuccessful landing (marked with a red $\times$). $\epsilon_N>\epsilon_\text{plat}$ increases time to land exponentially and increases mean absolute error as it hesitates to land.}
 \label{x_sim}
\end{figure}

\section{Simulation} \label{sim_sect}
We simulate a 2D multirotor with dynamics (\ref{multirotor}) in the $x-z$ plane landing on a cyclically heaving platform. The heave motion is modeled as $z_\text{plat}=A\sin(2\pi f t+\Phi)$ where $A$ is the amplitude in meters, $f$ is frequency in hertz, and $\Phi$ is phase offset in radians. Following Assumption \ref{a1}, we treat contact as a point force applied at the center of mass of the multirotor. In simulation, we are able to exaggerate and idealize the wave parameters to clearly demonstrate that the assumption of a perfectly inelastic collision ($\epsilon_N=0$) is invalid. For the simulation we select $A=0.1$ m, $f=1.5$ Hz, and $\Phi=\frac{5\pi}{8}$ rad.

To validate our proposed impact-aware LCP-MPC, we evaluate a series of MPC-based strategies:
1) track platform (red) $z_\text{ref}=z_\text{plat}$ with no LCP solve, 2) perfectly inelastic LCP-MPC (yellow) augmenting the track platform strategy with an LCP solve for the next-step normal impulse $\lambda_{N,k+1}^\star$ using $\epsilon_N=0$, and 3)-5) restitution-aware LCP-MPC augmenting the track platform strategy with an LCP solve using $\epsilon_N>0$. The restitution-aware cases are: 3) underdamped $\epsilon_N=0.25$ (black), 4) our proposed strategy $\epsilon_N=\epsilon_\text{plat}=0.5$ corresponding to the true platform restitution (blue), and 5) overdamped $\epsilon_N=0.75$ (green). Note that for the strategy without an LCP (red), the complementarity condition is not enforced in the dynamic equality constraints, we assume no knowledge of next-step normal impulse $\lambda_{N,k+1}^\star$, and residual cost $J_R(\cdot)=0$. In all cases the MPC is solved using CasADi (\cite{andersson_casadi_2019}) and the IPOPT software library.

In Fig. \ref{full_sim} we observe that the track position strategy (red), the perfectly inelastic LCP strategy with $\epsilon_N=0$ (yellow), and the low coefficient of restitution $\epsilon_N<\epsilon_\text{plat}$ strategy (black) strike the platform and rebound upwards due to the contact impulse. When the coefficient of restitution is underestimated relative to the true platform value, the UAV underestimates both the next-step normal impulse and the resulting post-impact velocity. Consequently, the MPC permits the UAV to approach the platform with high pre-impact velocity, producing unsuccessful rebound behaviour consistent with Newton's restitution law (\ref{eq:newton_method}). This effect is shown in Fig. \ref{partial_sim}, where the thrust commanded by the MPC to reduce the UAV's downward velocity is activated on average $0.02\,\text{ s}$ later and for $0.10\,\text{ s}$ less in the cases with $\epsilon_N<\epsilon_\text{plat}$ compared to those with $\epsilon_N\geq\epsilon_\text{plat}$. Our approach (blue) with $\epsilon_N=0.5$ achieves a $6.9\%$ decrease in mean absolute error (MAE) in $z$ between the UAV trajectory and the platform relative to the perfectly inelastic case ($\epsilon_N=0$). Underdamping ($\epsilon_N=0.25$) or overdamping ($\epsilon_N=0.75$) the coefficient of restitution by $0.25$ increases MAE by $4.0\%$ and $6.3\%$, respectively. Generally, as shown in Fig. \ref{x_sim}, deviations of $\epsilon_N$ from the true platform restitution value ($\epsilon_\text{plat}$) lead to an exponential increase in tracking error, highlighting restitution as a critical and tunable parameter for landings.

\section{Experiments}
\begin{figure}[t]
\centering
\includegraphics[width=0.325\textwidth,trim={20.0cm 8.5cm 14.0cm 2.3cm},clip]{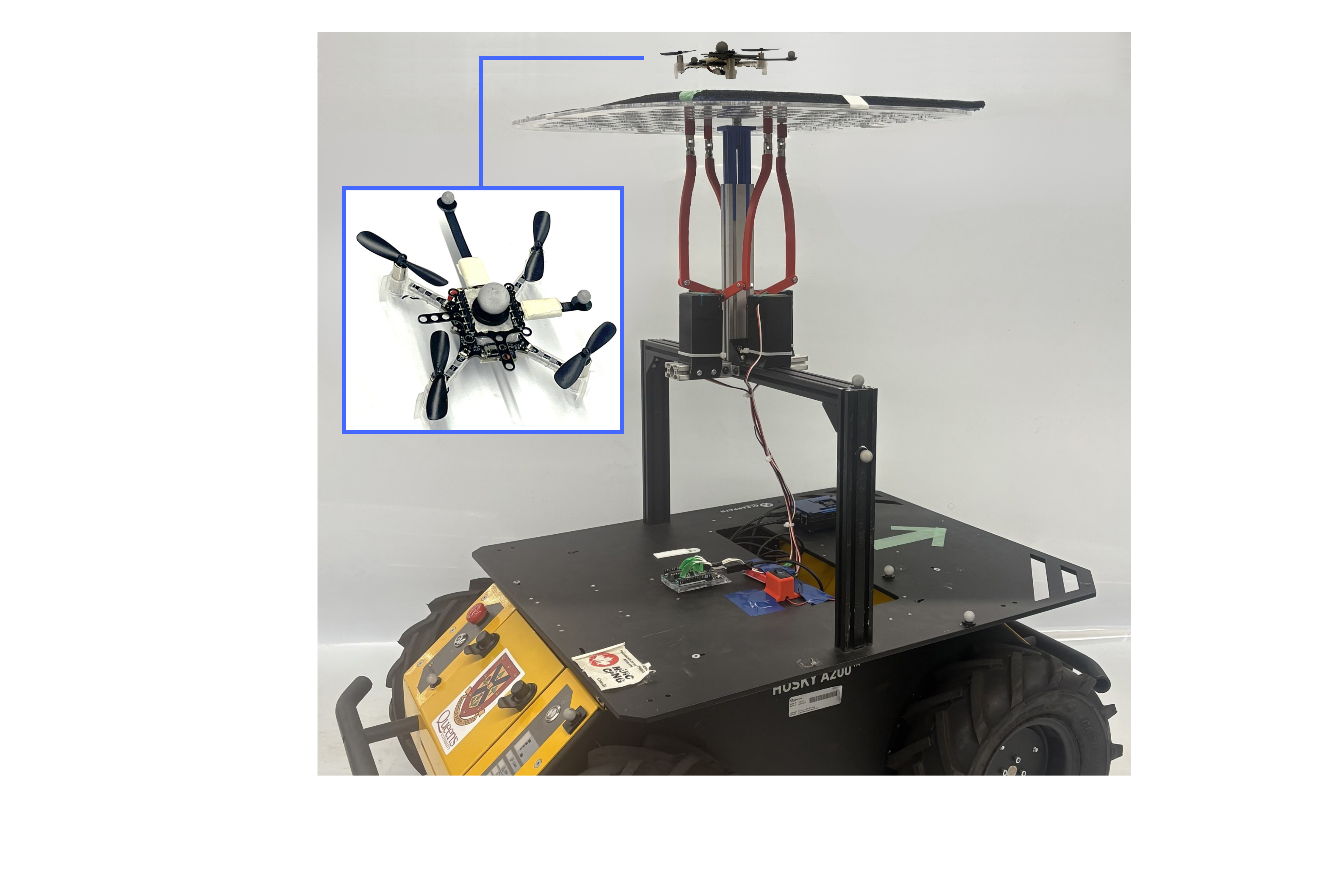}
 \caption{Experimental setup showing a Bitcraze Crazyflie 2.1 approaching the custom heaving deck emulating the motion of a USV heaving in waves. The platform is mounted on the ClearPath Robotics Husky unmanned ground vehicle (UGV).}
 \label{hardware}
\end{figure}

We evaluate the proposed restitution-aware, LCP-based MPC framework in indoor experiments comprising five trials per condition across four experimental conditions, generated by varying the amplitude and frequency of the platform's heave motion. Two strategies previously evaluated in simulation are compared: 1) baseline platform tracking (red) with $z_\text{ref}=z_\text{plat}$ and no LCP solve, and 2) our proposed restitution-aware strategy with $\epsilon_N=0.5$ corresponding to the estimated platform restitution (blue). The experimental setup, see Fig. \ref{hardware}, consists of a Bitcraze Crazyflie 2.1 UAV and a ClearPath Robotics Husky equipped with our custom heaving platform, a modified version of the tilting testbed, presented in \cite{stephenson_time}, enabling evaluation of landing strategies under a scaled approximation of real-world wave conditions. The heave motion is commanded according to the sinusoidal model used in simulation, restricted to amplitudes $0.01\leq A \leq0.03$ m, and frequencies $0.5\leq f\leq0.8$ Hz, which reflect the physical limitations of the hardware actuating the platform. Weight matrices in the tracking and restitution cost are selected as $\mathbf{Q}= 8\times 10^{6}\,\mathbf{I}$, $\mathbf{R}=10^{-3}\,\mathbf{I}$, and $W=0.1$. The LCP-MPC is run off-board on a Thinkpad X1 Carbon with Intel Core i7-1270P Processor at 10 Hz and transmits control inputs to the Crazyflie via long range Crazyradio USB at 100 Hz. The platform is stationary in $x$ and publishes its current $z$-position over a ROS topic on a local WiFi network. The Crazyflie receives its state feedback at 240 Hz from a VICON motion capture system; we assume that the UAV also has access to the platform's heave model for prediction, see Assumption \ref{a0}. 

\begin{figure}[t]
\centering
\includegraphics[width=0.9\linewidth,trim={0.4cm 0cm 0.37cm 0.4cm},clip]{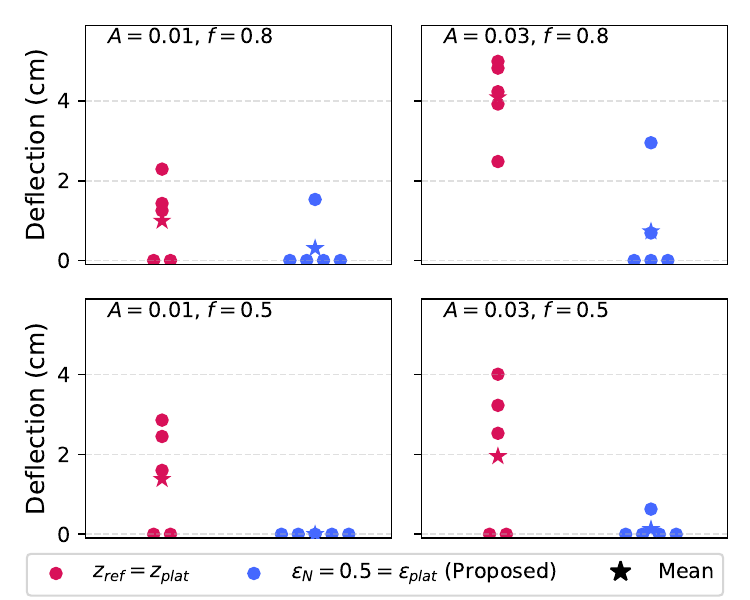}
\caption{Each point represents one of five trials at the indicated amplitude and frequency. The baseline tracking strategy (red) achieves a 30\% success rate with an average deflection height of $2.10$ cm, while the proposed strategy (blue) achieves an 80\% success rate with an average deflection height of $0.29$ cm.}
\label{dots}
\end{figure}

\begin{figure}[t]
\centering
\includegraphics[width=0.37\textwidth,trim={0.4cm 0.48cm 0.3cm 0.3cm},clip]{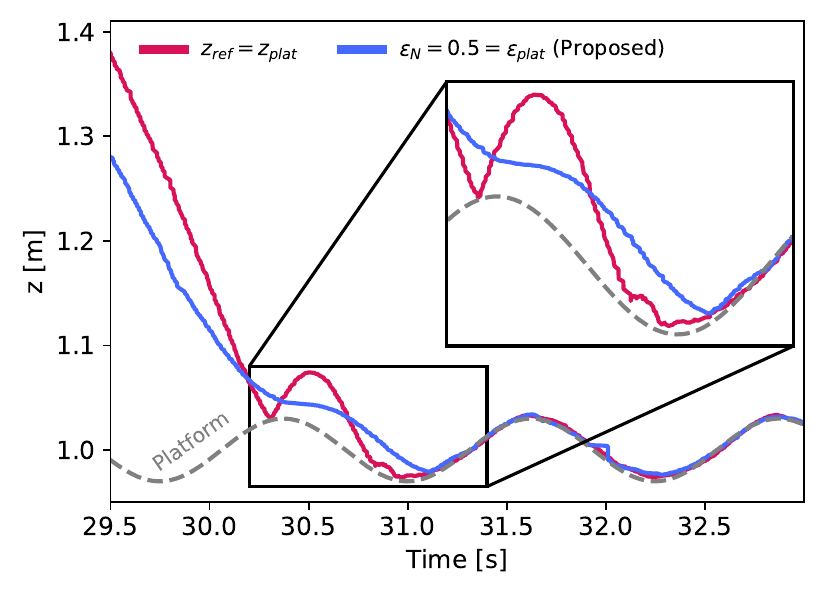}
 \caption{Landing trajectories from Trial 1 with $A=0.03$ m, $f=0.8$ Hz. The baseline tracking MPC (red) causes the UAV to impact the rising platform and rebound 4.23 cm due to high relative velocity. The proposed strategy (blue) with $\epsilon_N=0.5$ achieves zero deflection at landing but exhibits an overdamped response and a delayed touchdown.}
 \label{heave_real}
\end{figure}

We define an unsuccessful landing as any non-zero deflection of the UAV after first contact with the platform. Fig. \ref{dots} summarizes the post-impact deflection observed across all experimental trials, grouped by heave amplitude and frequency. In all four experimental conditions, the baseline tracking strategy (red) produces large rebounds, up to 4.99 cm, confirming the risk inherent in a pure tracking MPC that does not anticipate the effect of the impact on the UAV. In contrast, the proposed impact-aware LCP-MPC strategy substantially reduces post-impact deflection, achieving an 86.2\% reduction in average rebound height and a corresponding 166.7\% increase in landing success rate.
Fig. \ref{heave_real} shows the trajectories executed by the baseline and proposed MPC schemes for Trial 1 under the most severe wave conditions we evaluate (amplitude 0.03m, frequency 0.8 Hz). The baseline tracking strategy (red) exhibits the same behavior observed in simulation (Fig. \ref{full_sim}): the UAV collides with the platform as it is rising, generating a large pre-impact relative velocity and a resulting rebound of 4.23 cm (significant given the size and relatively small inertia of the Crazyflie UAV) due to the contact impulse. The proposed approach (blue) with $\epsilon_N=0.5$ achieves a successful landing with zero deflection, however, its trajectory resembles the overdamped behavior seen in Fig. \ref{sim4}, corresponding to a coefficient of restitution higher than the true platform value ($\epsilon_N=0.75>\epsilon_\text{plat}$). This overdamping produces an exaggerated reduction in pre-impact velocity and a delayed touchdown, up to approximately one second slower than the tracking baseline. These results indicate that the selected coefficient of restitution was overly conservative for this experimental platform. As suggested by the trends in Fig. \ref{x_sim}, deviations of $\epsilon_N$ in the LCP formulation from the true platform restitution ($\epsilon_\text{plat}$) lead to rapidly increasing time to land and subsequent larger tracking error as the UAV hesitates to land. This effect, compounded by actuation and communication delays in the physical system, likely contributed to the overdamped landing response.

\section{Future Work}
This work relies on several simplifying assumptions that enable controlled, reduced-scale validation; future efforts will remove these assumptions and transition toward realistic field trials on USVs in waves. First, we assume the platform’s vertical position is known, but future implementations will require sensing and estimation of deck motion with uncertainty. Second, we model landing as a single frictionless point contact; extending the framework to a multi-contact formulation with individual leg contacts would capture staggered touchdowns and impact-induced rotations. Third, the dynamics are restricted to 2D, and future work will incorporate full 6 degrees of freedom (DoF) UAV and USV motion.

Our indoor experiments also exclude wind and restrict the platform to pure heave, whereas real maritime conditions involve aerodynamic disturbances and complex 6-DoF vessel motion such as roll and pitch. We aim to evaluate the controller under these more realistic conditions in simulation and outdoor testing. Finally, we use an approximate, fixed restitution coefficient; future work will focus on online estimation or learning of restitution to improve performance across varying deck materials and landing conditions.

Formal stability guarantees for MPC with velocity-level complementarity constraints remain an open challenge. Because restitution induces discontinuous 
velocity jumps, standard Lyapunov-based MPC analyses do not directly apply (\cite{aydinoglu_contact-aware_2020}). Extending the framework with suitable hybrid stability certificates or 
nonsmooth Lyapunov tools is therefore an important direction for future work. In this paper, we instead evaluated robustness empirically through simulation 
and hardware experiments.

\section{Conclusion}
Heave motion poses a significant challenge for safe UAV landings on marine vessels, as the relative motion between a heaving platform and a descending vehicle can increase pre-impact relative velocity and produce large, uneven impact forces upon contact. In this paper, we propose an impact-aware, LCP-based MPC framework that explicitly models landing as an impact event. Our approach reduces post-impact deflection by 86.2\% over standard tracking MPCs across a range of wave conditions, enabling substantially safer landings.
\bibliography{ifacconf}

\begin{thebibliography}{31}
\providecommand{\natexlab}[1]{#1}
\providecommand{\url}[1]{\texttt{#1}}
\providecommand{\urlprefix}{URL }
\expandafter\ifx\csname urlstyle\endcsname\relax
  \providecommand{\doi}[1]{doi:\discretionary{}{}{}#1}\else
  \providecommand{\doi}{doi:\discretionary{}{}{}\begingroup \urlstyle{rm}\Url}\fi

\bibitem[{Abujoub et~al.(2020)Abujoub, McPhee, and Irani}]{abujoub_methodologies_2020}
Abujoub, S., McPhee, J., and Irani, R.A. (2020).
\newblock Methodologies for landing autonomous aerial vehicles on maritime vessels.
\newblock \emph{Aerospace Science and Technology}, 106.

\bibitem[{Alarcón et~al.(2019)Alarcón, García, Maza, Viguria, and Ollero}]{alarcon_precise}
Alarcón, F., García, M., Maza, I., Viguria, A., and Ollero, A. (2019).
\newblock A precise and gnss-free landing system on moving platforms for rotary-wing uavs.
\newblock \emph{Sensors}, 19(4).

\bibitem[{Andersson et~al.(2019)Andersson, Gillis, Horn, Rawlings, and Diehl}]{andersson_casadi_2019}
Andersson, J.A.E., Gillis, J., Horn, G., Rawlings, J.B., and Diehl, M. (2019).
\newblock {CasADi}: a software framework for nonlinear optimization and optimal control.
\newblock \emph{Mathematical Programming Computation}, 11(1), 1--36.

\bibitem[{Anitescu and Potra(1997)}]{anitescu1997formulating}
Anitescu, M. and Potra, F.A. (1997).
\newblock Formulating dynamic multi-rigid-body contact problems with friction as solvable linear complementarity problems.
\newblock \emph{Nonlinear Dynamics}, 14(3), 231--247.

\bibitem[{Aydinoglu et~al.(2020)Aydinoglu, Preciado, and Posa}]{aydinoglu_contact-aware_2020}
Aydinoglu, A., Preciado, V.M., and Posa, M. (2020).
\newblock Contact-{aware} {controller} {design} for {complementarity} {systems}.
\newblock In \emph{2020 {IEEE} {Int.} {Conf.} on {Rob.} and {Auto.} ({ICRA})}, 1525--1531.

\bibitem[{Banerjee et~al.(2016)Banerjee, Chanda, and Das}]{banerjee_oblique_2016}
Banerjee, A., Chanda, A., and Das, R. (2016).
\newblock Oblique frictional unilateral contacts perceived in curved bridges.
\newblock \emph{Nonlinear Dynamics}, 85(4), 2207--2231.

\bibitem[{Cottle et~al.(2009)Cottle, Pang, and Stone}]{cottle_linear_2009}
Cottle, R.W., Pang, J.S., and Stone, R.E. (2009).
\newblock \emph{The {Linear} {Complementarity} {Problem}}.
\newblock Society for Industrial and Applied Mathematics.

\bibitem[{Foehn et~al.(2017)Foehn, Falanga, Kuppuswamy, Tedrake, and Scaramuzza}]{foehn_fast_2017}
Foehn, P., Falanga, D., Kuppuswamy, N., Tedrake, R., and Scaramuzza, D. (2017).
\newblock Fast {trajectory} {optimization} for {agile} {quadrotor} {maneuvers} with a {cable}-{suspended} {payload}.
\newblock In \emph{Robotics: {Science} and {Systems} {XIII}}. Robotics: Science and Systems Foundation.

\bibitem[{Fuser et~al.(2025)Fuser, Nguyen, Incremona, Farina, and Cognetti}]{fuser_linear_2025}
Fuser, R., Nguyen, H.N., Incremona, G.P., Farina, M., and Cognetti, M. (2025).
\newblock A {linear} {complementarity} based {MPC} for {aerial} {physical} {interaction}.
\newblock In \emph{2025 {Int.} {Conf.} on {Unmanned} {Aircraft} {Systems} ({ICUAS})}, 982--987.

\bibitem[{Gupta et~al.(2022)Gupta, Pairet, Nascimento, and Saska}]{gupta2022landing}
Gupta, P.M., Pairet, E., Nascimento, T., and Saska, M. (2022).
\newblock Landing a uav in harsh winds and turbulent open waters.
\newblock \emph{IEEE Rob. and Auto. Letters}, 8(2), 744--751.

\bibitem[{Hervas et~al.(2014)Hervas, Reyhanoglu, and Tang}]{hervas_automatic_2014}
Hervas, J.R., Reyhanoglu, M., and Tang, H. (2014).
\newblock Automatic landing control of {unmanned} {aerial} {vehicles} on moving platforms.
\newblock In \emph{2014 {IEEE} 23rd {Int.} {Symp.} on {Industrial} {Electronics} ({ISIE})}, 69--74.

\bibitem[{Jean(1999)}]{jean_non-smooth_1999}
Jean, M. (1999).
\newblock The non-smooth contact dynamics method.
\newblock \emph{Computer Methods in Applied Mechanics and Engineering}, 177(3-4), 235--257.

\bibitem[{Le~Cleac'h et~al.(2024)Le~Cleac'h, Howell, Yang, Lee, Zhang, Bishop, Schwager, and Manchester}]{lecleach}
Le~Cleac'h, S., Howell, T.A., Yang, S., Lee, C.Y., Zhang, J., Bishop, A., Schwager, M., and Manchester, Z. (2024).
\newblock Fast contact-implicit model predictive control.
\newblock \emph{IEEE Transactions on Robotics}, 40, 1617--1629.

\bibitem[{Lemke(1954)}]{lemke_dual_1954}
Lemke, C.E. (1954).
\newblock The dual method of solving the linear programming problem.
\newblock \emph{Naval Research Logistics Quarterly}, 1(1), 36--47.

\bibitem[{León et~al.(2021)León, Rimoli, and Di~Leo}]{leon_rotorcraft_2021}
León, B.L., Rimoli, J.J., and Di~Leo, C.V. (2021).
\newblock Rotorcraft {dynamic} {platform} {landings} {using} {robotic} {landing} {gear}.
\newblock \emph{Journal of Dynamic Systems, Measurement, and Control}, 143(11), 111006.

\bibitem[{Lyu et~al.(2023)Lyu, Zhao, Huang, and Huang}]{lyu_unmanned_2023}
Lyu, M., Zhao, Y., Huang, C., and Huang, H. (2023).
\newblock Unmanned {aerial} {vehicles} for {search} and {rescue}: {a} {survey}.
\newblock \emph{Remote Sensing}, 15(13), 3266.

\bibitem[{Marconi et~al.(2002)Marconi, Isidori, and Serrani}]{marconi_autonomous_2002}
Marconi, L., Isidori, A., and Serrani, A. (2002).
\newblock Autonomous vertical landing on an oscillating platform: an internal-model based approach.
\newblock \emph{Automatica}, 38(1), 21--32.

\bibitem[{Moreau(1988)}]{moreau_unilateral_1988}
Moreau, J.J. (1988).
\newblock Unilateral {contact} and {dry} {friction} in {finite} {freedom} {dynamics}.
\newblock In J.J. Moreau and P.D. Panagiotopoulos (eds.), \emph{Nonsmooth {Mechanics} and {Applications}}, volume 302, 1--82. Springer Vienna.

\bibitem[{Persson(2021)}]{persson2021model}
Persson, L. (2021).
\newblock \emph{Model Predictive Control for Cooperative Rendezvous of Autonomous Unmanned Vehicles}.
\newblock Ph.D. thesis, KTH Royal Institute of Technology.

\bibitem[{Posa et~al.(2014)Posa, Cantu, and Tedrake}]{posa_direct_2014}
Posa, M., Cantu, C., and Tedrake, R. (2014).
\newblock A direct method for trajectory optimization of rigid bodies through contact.
\newblock \emph{The Int. Journal of Robotics Research}, 33(1), 69--81.

\bibitem[{Ross et~al.(2021)Ross, Seto, and Johnston}]{ross_autonomous_2021}
Ross, J., Seto, M., and Johnston, C. (2021).
\newblock Autonomous {landing} of {rotary} {wing} {unmanned} {aerial} {vehicles} on {underway} {ships} in a {sea} {state}.
\newblock \emph{Journal of Intelligent \& Robotic Systems}, 104(1).

\bibitem[{Ross et~al.(2019)Ross, Seto, and Johnston}]{ross_zero_2019}
Ross, J., Seto, M.L., and Johnston, C. (2019).
\newblock Zero visibility autonomous landing of quadrotors on underway ships in a sea state.
\newblock In \emph{2019 IEEE Int. Symp. on Robotic and Sensors Environments (ROSE)}, 1--7.

\bibitem[{Sanchez-Lopez et~al.(2014)Sanchez-Lopez, Pestana, Saripalli, and Campoy}]{sanchez-lopez_approach_2014}
Sanchez-Lopez, J.L., Pestana, J., Saripalli, S., and Campoy, P. (2014).
\newblock An {approach} {toward} {visual} {autonomous} {ship} {board} {landing} of a {VTOL} {UAV}.
\newblock \emph{Journal of Intelligent \& Robotic Systems}, 74(1), 113--127.

\bibitem[{Signorini(1933)}]{signorini1933sopra}
Signorini, A. (1933).
\newblock Sopra alcune questioni di elastostatica.
\newblock \emph{Atti della Societa Italiana per il Progresso delle Scienze}, 21(2), 143--148.

\bibitem[{Stephenson et~al.(2025)Stephenson, Stewart, and Greeff}]{stephenson_time}
Stephenson, J., Stewart, W.S., and Greeff, M. (2025).
\newblock A time and place to land: Online learning-based distributed {MPC} for multirotor landing on surface vessel in waves.
\newblock In \emph{2025 Int. Conf. on Unmanned Aircraft Systems (ICUAS)}, 193--199.

\bibitem[{Stewart and Trinkle(2000)}]{stewart2000implicit}
Stewart, D. and Trinkle, J.C. (2000).
\newblock An implicit time-stepping scheme for rigid body dynamics with coulomb friction.
\newblock In \emph{IEEE Int. Conf. on Rob. and Auto. (ICRA). Symposia Proceedings}, volume~1, 162--169.

\bibitem[{Van~Bokhoven and Leenaerts(1999)}]{van_bokhoven_explicit_1999}
Van~Bokhoven, W. and Leenaerts, D. (1999).
\newblock Explicit formulas for the solutions of piecewise linear networks.
\newblock \emph{IEEE Transactions on Circuits and Systems I: Fundamental Theory and Applications}, 46(9), 1110--1117.

\bibitem[{Vlantis et~al.(2015)Vlantis, Marantos, Bechlioulis, and Kyriakopoulos}]{Panagiotis2015}
Vlantis, P., Marantos, P., Bechlioulis, C.P., and Kyriakopoulos, K.J. (2015).
\newblock Quadrotor landing on an inclined platform of a moving ground vehicle.
\newblock In \emph{2015 IEEE Int. Conf. on Rob. and Auto. (ICRA)}, 2202--2207.

\bibitem[{Xia et~al.(2022)Xia, Shin, Chung, Kim, Lee, and Son}]{xia2022landing}
Xia, K., Shin, M., Chung, W., Kim, M., Lee, S., and Son, H. (2022).
\newblock Landing a quadrotor uav on a moving platform with sway motion using robust control.
\newblock \emph{Control Engineering Practice}, 128.

\bibitem[{Xuan-Mung et~al.(2020)Xuan-Mung, Hong, Nguyen, Ha, and Le}]{xuan-mung_autonomous_2020}
Xuan-Mung, N., Hong, S.K., Nguyen, N.P., Ha, L.N.N.T., and Le, T.L. (2020).
\newblock Autonomous {quadcopter} {precision} {landing} {onto} a {heaving} {platform}: {new} {method} and {experiment}.
\newblock \emph{IEEE Access}, 8, 167192--167202.

\bibitem[{Yeong et~al.(2015)Yeong, King, and Dol}]{yeong_review_2015}
Yeong, S.P., King, L.M., and Dol, S.S. (2015).
\newblock A {review} on {marine} {search} and {rescue} {operations} {using} {unmanned} {aerial} {vehicles}.
\newblock \emph{Int. Journal of Marine and Environmental Sciences}, 9(2), 396--399.

\end{thebibliography}
\end{document}